# Water quality polluted by total suspended solids classified within an Artificial Neural Network approach

I. Luviano Soto[a]*, Y. Concha Sánchez[a], and A. Raya[b,c]

[a] Facultad de Ingeniería Civil, Universidad Michoacana de San Nicolás de Hidalgo. Edificio C, Ciudad Universitaria. Francisco J. Mújica S/N. Col. Felícitas del Río. C.P. 58030, Morelia, Michoacán, México.

[b] Instituto de Física y Matemáticas, Universidad Michoacana de San Nicolás de Hidalgo. Edificio C-3, Ciudad Universitaria. Francisco J. Mújica S/N, Col. Felícitas del Río. C.P. 58040, Morelia, Michoacán, México.

[c] Centro de Ciencias Exactas, Universidad del Bío-Bío. Casilla 447, Chillán, Chile.

*Corresponding author. E-mail: itzel.luviano@umich.mx

ABSTRACT

This study investigates the application of an artificial neural network framework for analysing water pollution caused by solids. Water pollution by suspended solids poses significant environmental and health risks. Traditional methods for assessing and predicting pollution levels are often time-consuming and resource-intensive. To address these challenges, we developed a model that leverages a comprehensive dataset of water quality from total suspended solids. A convolutional neural network was trained under a transfer learning approach using data corresponding to different total suspended solids concentrations, with the goal of accurately predicting low, medium and high pollution levels based on various input variables. Our model demonstrated high predictive accuracy, outperforming conventional statistical methods in terms of both speed and reliability. The results suggest that the artificial neural network framework can serve as an effective tool for real-time monitoring and management of water pollution, facilitating proactive decision-making and policy formulation. This approach not only enhances our understanding of pollution dynamics but also underscores the potential of machine learning techniques in environmental science.

**Keywords:** Artificial Neural Networks, Total Suspended Solids, Water Pollution.

**HIGHLIGHTS**

- Artificial neural networks are used to study water pollution by total suspended solids.
- Samples are prepared with different concentrations of suspended solids and filmed to obtain images
- Convolutional neural network with a transfer learning strategy is trained and tested for performance.

- Low, medium and high concentrations of total suspended solids in water can be identified with high accuracy and modest computer resources.

**GRAPHICAL ABSTRACT**

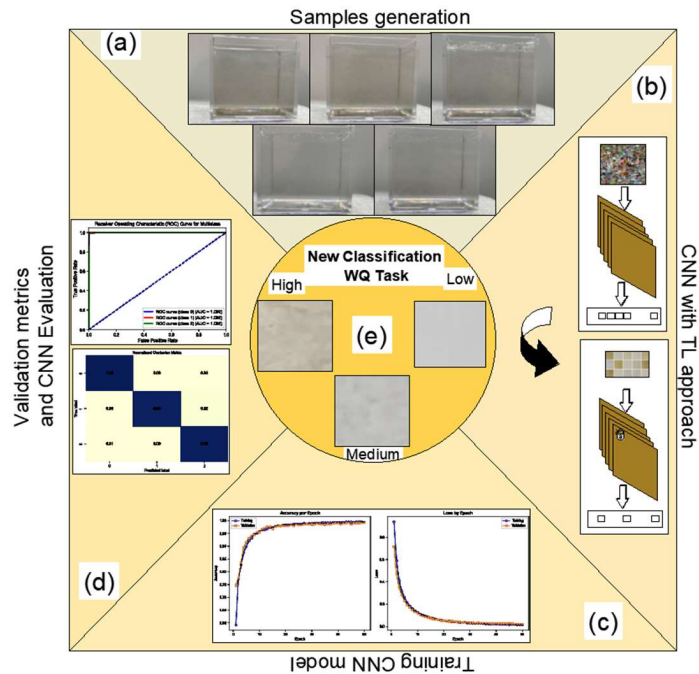

**INTRODUCTION**

Water pollution by solids, including suspended and dissolved particles, poses critical challenges to environmental sustainability, public health, and economic development (Khan *et al*. 2022). Understanding and mitigating these impacts require accurate monitoring and prediction of pollution levels. For instance, in Mexico, the Ministry of Environment and Natural Resources has proposed NOM-001-SEMARNAT-2021 (National Advisory Committee for Standardization of the Environment and Natural Resources of Mexico 2021) as the standard for setting parameters for contaminated water quality, including Total Suspended Solids

(TSS). To make water quality information accessible to the public, the Ministry also proposes a system for estimating water quality by measuring three indicative parameters that differentiate various water quality levels. TSS is among these parameters because elevated TSS levels diminish the ability of water bodies to support diverse aquatic life. These parameters help to identify conditions ranging from nearly natural, unaffected by human activity, to water showing clear signs of significant municipal and non-municipal wastewater discharges and severe deforestation (Ministry of Environment and Natural Resources Mexico 2011).

Traditional methods to identify and classify solid pollutants in water samples include Gooch crucible with 2.4 cm glass fibre filter, Buchner funnel, membrane filter, Gooch crucible asbestos (Smith *et al.* 1963). While effective, these methods often involve labour-intensive sampling and analysis processes that can be both time-consuming and costly. This underscores the need for innovative approaches to enhance the efficiency and accuracy of water quality assessment. An Artificial Neural Network (ANN) framework offers a promising solution to these challenges. To mention a few features in which an ANN framework is a valuable tool, we should keep in mind that water quality data is inherently complex, with numerous interdependent variables such as turbidity, TSS, and chemical oxygen demand (COD) (Tchobanoglous *et al.* 1991), for which an ANN excel in managing and interpreting complex, non-linear relationships within large datasets, providing more accurate predictions than traditional linear models. Such a framework is capable of learning from historical data to predict future pollution levels, which is crucial for proactive environmental management (Schauser & Steinberg 2001). Furthermore, once trained, ANN models can rapidly process and analyse data, significantly reducing the time required for water quality assessment, a beneficial feature for real-time monitoring and decision-making (Palani *et al.* 2008). ANN frameworks can be easily scaled to incorporate additional data sources and parameters, enhancing their applicability across different water bodies and pollution scenarios. These frameworks continuously improve their performance as more data becomes available and thus remain relevant and accurate over time, adapting to changes in pollution patterns and environmental conditions. An additional benefit of the use of ANNs is that these frameworks can be seamlessly integrated with Internet of Things (IoT) devices and remote sensing technologies. This integration enables continuous, automated monitoring of water quality, providing real-time data that enhances the responsiveness and effectiveness of pollution control measures. By providing reliable and detailed insights into pollution dynamics, ANN frameworks support informed decision-making by policymakers, environmental agencies, and stakeholders. This leads to better resource allocation, targeted pollution control strategies, and ultimately, more effective environmental protection.

In the field of water management, several ANN models have been primarily used to characterise both the quantity and quality of water (Farmaki *et al.* 2010). One of the most common applications of ANNs in water monitoring is through remote sensing (Wagle & Acharya 2020), where satellite images are used to predict different water levels and the evolution of contaminants (Agrawal & Peterson 2021). Another significant application is in predicting wastewater, particularly in Wastewater Treatment Plants (WWTPs), as well as in process control within these facilities (Hamed *et al.* 2004). Also, ANNs are widely utilised for predicting water quality parameters through various machine learning methods (Haghiabi *et al.* 2018).

In this article, we present a novel approach to assess water quality by developing a Convolutional Neural Network (CNN) capable of predicting high, medium, and low pollution levels based on TSS. This method offers a cost-effective, rapid, and non-invasive alternative for water quality monitoring, enabling the determination of water quality categories ('high', 'medium' and 'low') based on TSS concentration using a single image captured with a smartphone camera. Early studies under similar conditions already point toward the benefits of using CNN in classification of water contamination by TSS

(Lopez-Betancur *et al.* 2022). Our CNN, consisting of five layers, has been designed to detect TSS concentrations ranging from 40 to 6000 mg/L and provides the corresponding classifications. The remaining of this research work is organised as follows: In the Methods Section, we present in detail the sample preparation, the experimental procedure, the CNN development for classifying water quality based on solids, as well as its training and calibration phases. In the Results Section we report the findings of our study based upon the followed methodology. In the Discussion Section we summarise the findings of this work and its impact in environmental science. Finally, we present an outlook for future work in the Conclusions Section.

 **METHODS**

*Samples*

To carry out this research, 30 water samples were prepared with different levels of solids as main pollutant, this samples were obtained by selection clays with particle diameter size smaller than 60 micrometres since TSS are considered matter with particle diameter less than 62 micrometres (Bilotta & Brazier 2008). This process involved sieving the material through a 60-micron mesh sieve to achieve material homogenization. It is important to note that the selected clays are primarily composed of iron and aluminium, which are the most common types of clays found in urban environments, aimed at mimicking natural contamination in urban water sources (Perry & Taylor 2009). The water used for preparing the samples was distilled water, known for its purity, ensuring that TSS were the primary pollutant in the samples. The concentration range of the samples used was from 40 to 6000 mg/L, obtained by mixing pure water with clays. The clays were weighed on an analytical balance with a precision of 0.1 mg and divided into three categories of water polluted by solids: low, medium, and high. For low quality water were considered samples with concentrations from 40 to 70 mg/L. Water classified as medium quality by TSS contaminants ranges from 80 to 400 mg/L, while high quality water encompasses concentrations from 500 to 6000 mg/L. The number of samples varied across different classes: four samples for the low class, ten samples for the medium class, and sixteen samples for the high class. This variation is due to the differing ranges of TSS concentrations within each class. The low class exhibited minimal variability in TSS concentrations, the medium class displayed a broader range of concentrations, and the high class had the widest range of TSS concentrations.

*Experimental procedure*

To carry out the experimental part of this work, it was necessary to take photographic records of each water sample containing solids. This record was made by placing 100 mL of the sample in transparent cubic containers with a 5 cm edge. Each sample was illuminated laterally using a white dispersed light source, specifically an 18-inch Ring Light Edge-Lit LED, positioned 20 centimeters from the water sample. The light was placed laterally to avoid reflections on the water container. Due to the particle size of the TSS, a magnetic stirrer was required to prevent sedimentation. The stirrer operated at 300 rpm, which was the speed at which the solids remained in constant suspension, with a 1.5-inch diameter hexagonal capsule placed at the center of the sample. The images were captured through a 1.0-minute video recorded with an iPhone 12 with a 12 MP camera and a resolution of 1920 x 1080 progressively displayed pixels, also known as High Definition (HD), capable of 30 fps (frames per second) and a 2.5x zoom. The entire experimental setup was conducted on a levelled anti-vibration optical table, with the smartphone mounted statically on a tripod and positioned 15 centimeters from the camera shutter, located in a booth with black curtains to prevent external light intrusion. The entire experimental setup was documented to ensure repeatability in video recording. The images used in the development of the CNN were extracted from videos captured at a rate of 4 frames per second (fps). This process was implemented using Python, a versatile and accessible programming language (Thaker & Shukla 2020), with the MoviePy package. A total of 240 images were generated per sample, resulting in 7200 images. Following individual analysis, 685 images were excluded due to blur issues, resulting in a dataset of 6515 images for further analysis. The original dimensions of the captured images were 1920 x 1080 pixels, but they were centrally cropped to 450 x 450 pixels to reduce noise from the magnetic stir bar, container edges, and induced vortices by the magnetic stirrer. The cropping process was conducted using the OpenCV library in Python.

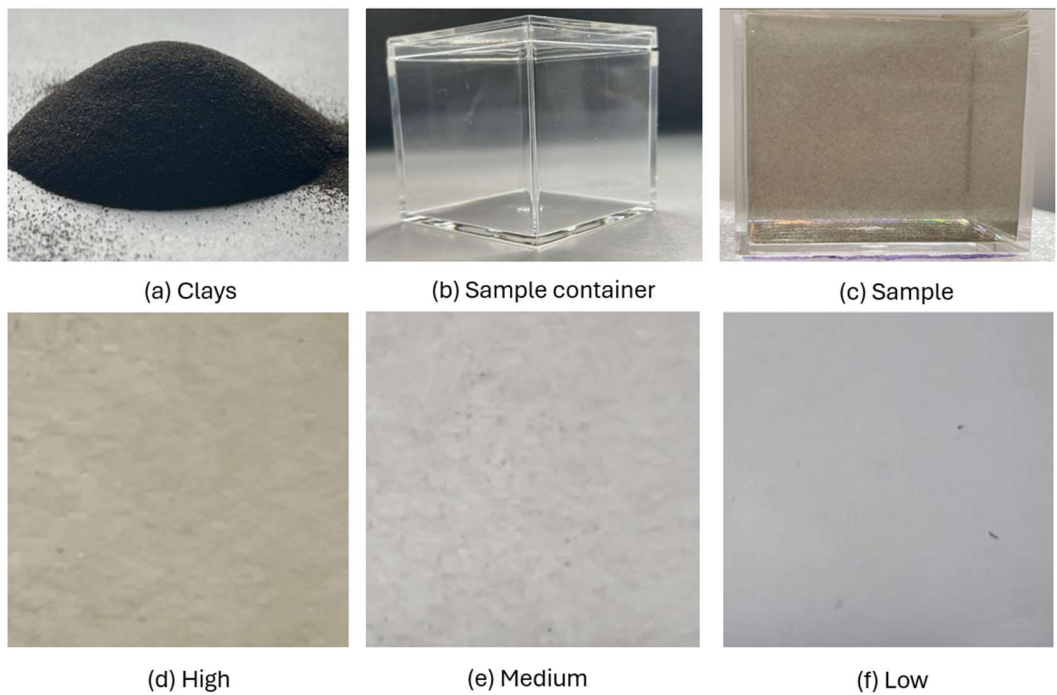

Figure 1. Samples supplies and data image obtained for the cropped image from the video recording: (a) Clays used to make the samples; (b) Container of 5 x 5 x 5 cm³; (c) A sample with the translucent container; (d) Cropped image associate a "high" TSS concentration; (e) Cropped image associate a "medium" TSS concentration; (f) Cropped image associate a "low" TSS concentration.

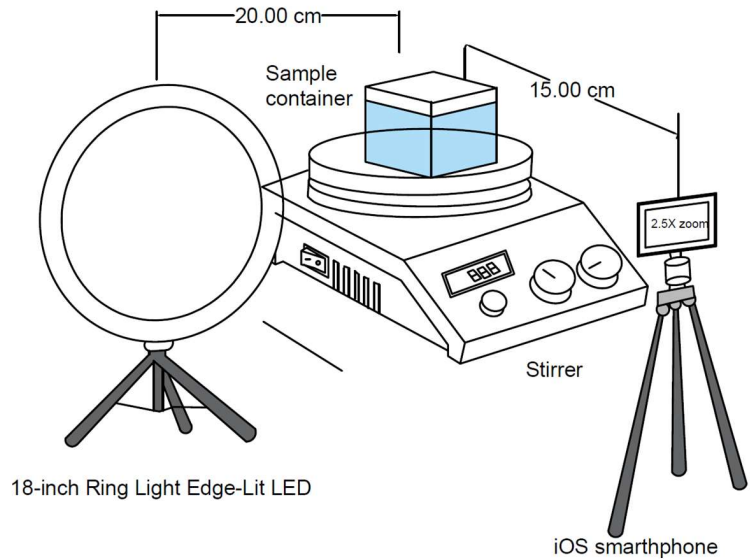

Figure 2. Scheme of experimental setup (see text). Samples are put in a transparent cubic container on a magnetic stirred, illuminated laterally using an 18-inch Ring Light Edge-Lit LED. Images were captured with an iPhone 12.

*CNN development*

In the domain of Deep Learning (DL), various types of algorithms exist, with Convolutional Neural Networks (CNNs) being among the most widely used (Baek *et al.* 2020). These models offer several advantages: (1) They reduce both time and costs (e.g., material and labour costs), (2) They enable forecasting across different system phases, (3) They simplify complex systems to enhance comprehension, and (4) They predict target values even in situations where site access is challenging (Barzegar *et al.* 2020). Therefore, in this study, we propose the use of a CNN.

The primary task of a CNN is classification. Initially, it performs feature extraction from the input image. These features are then fed into a Neural

Network (NN), producing output probabilities that indicate the classification of the input image into a specific category (Ferentinos 2018). However, training a CNN from scratch requires two main conditions: (1) Access to a large dataset with labelled data, and (2) Significant computational and memory resources (Morid *et al.* 2021).

An alternative to training CNNs from scratch is Transfer Learning (TL), which allows leveraging knowledge acquired from large datasets of non-environmental data to address specific environmental challenges, such as water quality analysis. Specifically, parameters from well-trained CNN models on non-environmental datasets, which contain diverse images (e.g., ImageNet models like AlexNet (Yuan & Zhang 2016), VGGNet (Purwono *et al*. 2023), and ResNet (Wu *et al.* 2019.)), can be transferred to tailor a CNN model for analysing water quality.

The use of TL with AlexNet is well known, as it has been used in various areas. For example, it has been used in the detection of pathologies (Lu *et al*. 2019), identification of alcoholism (Wang *et al.* 2019), multiple sclerosis (Zhang *et al.* 2019), skin lesions (Hosny *et al.* 2020), and even facial emotions (Sekaran *et al.* 2021). In the environmental area, AlexNet has also been used in the identification of diseases in maize leaves (Lv *et al.* 2020), identification of crop water stress (Chandel *et al.* 2019), sounds of marine mammals (Lu *et al.* 2021). However, in water quality classification, it has been little explored. In this study, we use TL for the classification of TSS using the AlexNet network, which is based on a CNN model that won the ImageNet Large Scale Visual Recognition Challenge 2012 (ILSVRC2012). This competition has served as a major benchmark for image recognition since 2010 (Russakovsky *et al.* 2015). AlexNet leverages the ImageNet dataset, which contains over 15 million labelled images (Deng, *et al.* 2009). Its architecture includes eight learned layers: five convolutional layers followed by Max-pooling layers to reduce data dimensions, utilising Rectified Linear Units (ReLu) as activation function, and three fully connected layers (Krizhevsky *et al.* 2017). In Figure 3 we show the architecture used in AlexNet CNN.

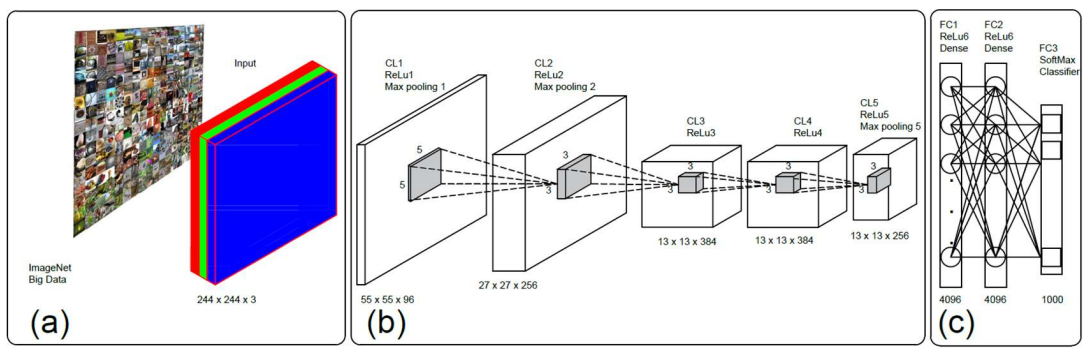

Figure 3. AlexNet architecture. (a) Training with big data ImageNet (input image size 244 x 244 x 3); (b) Five convolutional layers with ReLu as activation Function and Max-pooling to size reduce; (c) Three Fully Connected Layers (FC) and the last one layer uses SoftMax as activation Function to the classifier.

To carry out the TL process used in our work, we make use of Python as the programming language, because it has a large number of packages that facilitate the use of Machine Learning (ML) algorithms (Raschka & Mirjalili, 2019), and it is also free to use. The entire procedure was conducted using the Jupyter notebook environment within the Integrated Development Environment (IDE) Visual Studio Code. All the computations were performed on a computer equipped with an 8th-generation Intel i5 processor and 8 GB of RAM, which provides relatively modest computational resources. The well-trained AlexNet model was obtained from the Torchvision package, which is a Python package that includes several pre-trained network models. The modification of the AlexNet CNN involved removing the classifier, which initially had 1000 different types of classes, and we proposed a three-class water quality classifier based on different concentrations of TSS: High (500 to 6000 mg/L), Medium (80 to 400 mg/L), and Low (40 to 70 mg/L). Subsequently, the parameters obtained from the pre-training of AlexNet were unfrozen to adjust the weights of the network model.

In the ML realm, one of the most important elements to define are the hyperparameters of the ANN. These consist of configurations of the network that affect its structure, learning, and performance. Hyperparameters differ from parameters in that they are not automatically modified or adjusted during training; instead, they must be specified beforehand (Yu & Zhu 2020). The hyperparameters established in this work were: the optimization algorithm, learning rate, batch size, and number of epochs. Below, we describe each of these.

The optimization process in Artificial Intelligence (AI) involves identifying optimal parameters that improve the performance of a CNN model. One of the classic methods for this process is the Stochastic Gradient Descent (SGD) optimization method (Newton *et al.* 2018). However, tuning the learning rate of SGD, as a hyperparameter, is often challenging because the magnitudes of different parameters vary significantly and need to be adjusted throughout the training process (Zhang 2018). Therefore, in our study, we utilised the Adam optimizer, as it is an efficient stochastic optimization method that only requires first-order gradients and has low memory requirements. This method computes individual adaptive learning rates for different parameters based on estimates of the first and second moments of gradients (Kingma & Ba 2014), and iteratively finding values that minimise the error (loss).

In a deep NN, the weight parameters $\theta$ are updated as

$$\theta^t = \theta^{t-1} - \epsilon t \frac{\partial L}{\partial \theta} \qquad (1)$$

where $L$ represents the loss function and $\epsilon t$ the learning rate. Regarding the latter, it is known that a low rate causes slow convergence of the training algorithm, while a very high rate can lead to divergence (Smith 2017). Therefore, in this work, we chose to use a low learning rate to ensure model convergence, opting for a learning rate Lr=0.000005, at the expense of sacrificing some training convergence speed.

One of the crucial hyperparameters is the number of epochs used during training. An epoch entails presenting each sample in the training dataset with an opportunity to update the model internal parameters. Each epoch consists of one or more batches. As each sample or batch is processed through the network, the error is computed, and the Back Propagation (BP) algorithm is applied to adjust the weights and biases of the network. During BP, the error is propagated backward through the network, gradients of the weights are calculated with respect to the error, and these gradients are used to update the weights, aiming to minimise the error. This process facilitates model learning and performance enhancement. Although the number of epochs is typically large, our approach employs 50 epochs.

Conversely, the batch size determines the number of samples propagated through the CNN and used to update model parameters in each iteration until training is complete. Larger batch sizes facilitate greater computational parallelism and can often enhance performance. However, they require more memory and can introduce latency during training. For the creation of the CNN model, a total of 6,515 images with varying suspended solids concentrations were used, with 5,862 images allocated for the training set and 653 images for validation. Consequently, given the total number of images available, we opted for a batch size of 50 images.

The hyperparameters were chosen through numerous performance evaluations aimed at minimising error and maximising classification accuracy, specifically tailored to our image dataset. Table 1 presents the hyperparameters utilised for CNN development, while Figure 4 depicts the TL processes employed.

Table 1. Hyperparameters used in the development of the CNN model.

| Hyperparameters | Value |
|---|---|
| Algorithm Optimizer | Adam |
| Learning rate | 0.000005 |
| Batch Size | 50 |
| Epoch | 50 |

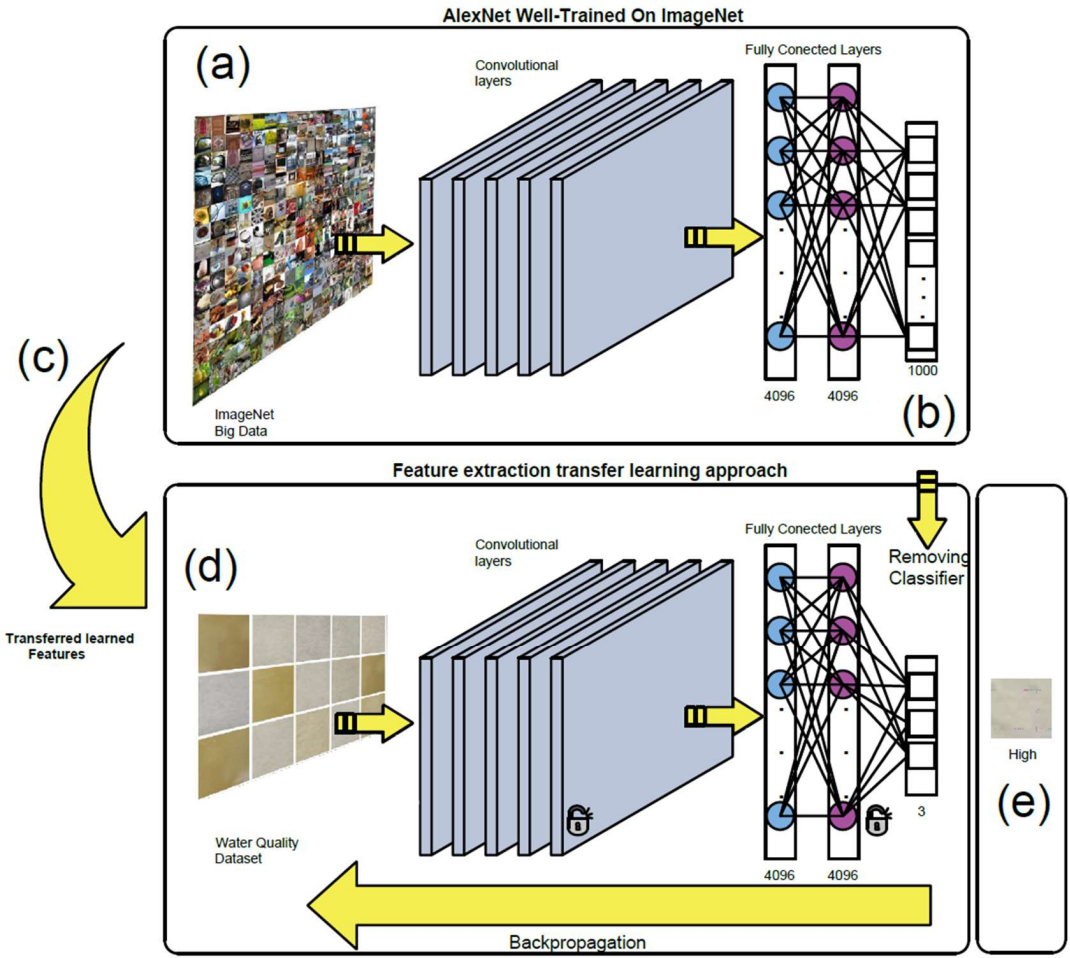

Figure 4. Representation of the TL process in the CNN model used: (a) Well-trained AlexNet CNN; (b) Removing last layer classifier for a new task; (c) Reuse of the pre-trained model; (d) New training dataset of water samples and modification of the CNN; (e) Classification result.

*Validation Metrics*

For evaluating the proposed CNN, the most common metrics are based on the prediction of four possible outcomes: True Positives ($TP$), True Negatives ($TN$), False Positives ($FP$), and False Negatives ($FN$) (Seliya *et al.* 2009). In this study, we use Accuracy, Precision, Recall, F-measure, Receiver Operating

Characteristic (ROC), and Confusion Matrix as validation metrics. Below, we describe the function of each metric.

Accuracy ($ACR$) represents the Classification Accuracy Rate at the decision, and is defined as

$$ACR = \frac{TP+TN}{N} \qquad (2)$$

where $N$ represents the total number of instances in the dataset.

Precision, defined in terms of the Positive Predictive Value ($PPV$) is obtained as

$$PPV = \frac{TP}{TP+FP} \qquad (3)$$

Sensitivity metric is represented by Recall, which indicates the true positive rate and measures the ability of the classifier to correctly identify positive instances. Sensitivity, or Recall, is defined as

$$Recall = \frac{TP}{TP+FN} \qquad (4)$$

F-measure (FM) or F-score metric is derived from two parameters, Recall and Precision (Witten *et al.* 2005). This measure, ranging from 0 to 1, peaks at 1 for a perfect classifier. F-measure is obtained as

$$F-measure = \frac{(1+\beta^2) \times Recall \times Precision}{Recall + Precision} \qquad (5)$$

In our study, $\beta$=1.

The ROC curve illustrates how a classifier balances between correctly identifying $TP$ and $FP$. It provides a comprehensive view of the classifier effectiveness, independent of class distribution or error costs (Davis & Goadrich 2006). The Area Under the ROC Curve (AUC) represents the probability that a randomly selected positive instance is ranked higher than a randomly selected negative instance according to the model predictions. Generally, a classifier with a larger AUC indicates better performance compared to one with a smaller AUC. This curve is commonly used as a validation metric.

Finally, a Confusion Matrix, a widely used tool in classification problems, is employed in this research. This tool provides detailed information about the predicted classifications (Deng *et al.* 2016). The Confusion Matrix is particularly beneficial for evaluating the overall performance of the classification model, which is crucial for guiding subsequent improvements. It is structured such that each cell corresponds to a specific class assigned by the model, with rows representing the actual classes and columns representing the predicted classes. Ideally, correctly classified instances align along the diagonal of the matrix, while misclassified instances appear in the off-diagonal cells.

**RESULTS**

The development of the CNN training process and its validation results are depicted in Figure 5, which illustrates the accuracy and loss curves. Figure

5a depicts the accuracy obtained for both the training and validation data as a function of time (epochs). Figure 5b also shows the error performance of the model on the y-axis, alongside the loss for each epoch. Our model initially achieved an accuracy of 0.78, but starting from epoch 10, it reached a precision of 0.99, with the loss minimised to approximately 0.005, varying between 0.02 and 0.0004 in each subsequent epoch. By epoch 50, the model attained an accuracy of 0.9985.

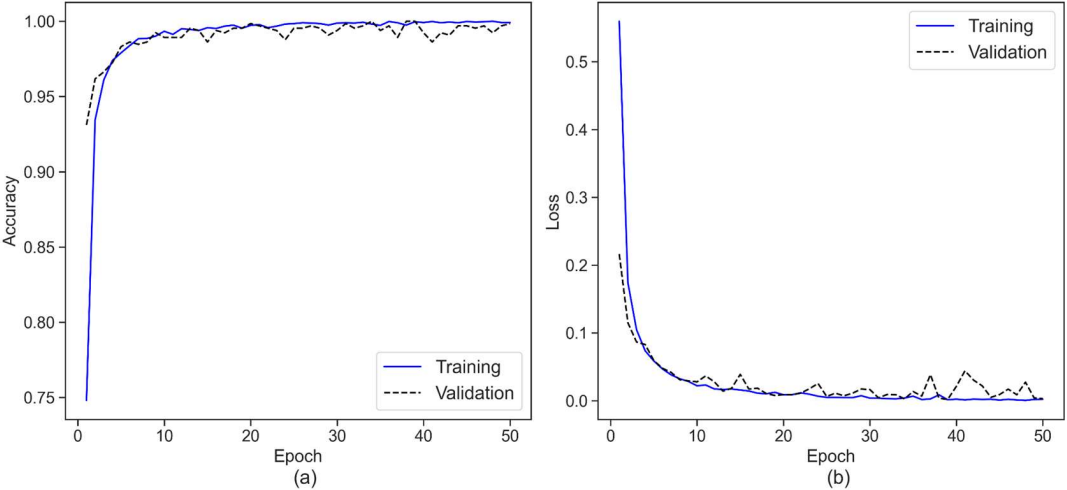

Figure 5. Learning curve of the CNN. (a) Classifier accuracy by epoch for training and validation dataset. (b) Classifier loss by epoch for training and validation dataset.

Figure 6 displays the Confusion Matrix obtained at the end of the training process for the three proposed classes. Both the normalised (Figure 6a) and unnormalised matrices (Figure 6b) are presented, based on 653 validation images: 360 for the "high" class, 206 for the "medium" class, and 87 for the "low" class. In the Confusion Matrix rows correspond to the true labels, whereas columns represent the labels predicted by the CNN proposed. Cells along the diagonal are associated with observations that have been correctly classified. Figure 6 indicates that the model achieves nearly 100% accuracy in label prediction, with minimal errors, misclassifying only one image associated with the "medium" label. The "high" class, which had the most images used during training, exhibited the best prediction accuracy, with all 360 validation images correctly classified by the CNN model. For the normalised Confusion Matrix (Figure 6a), two decimal places were used in rounding, showing a prediction accuracy of 100%, despite the presence of a misclassified image.

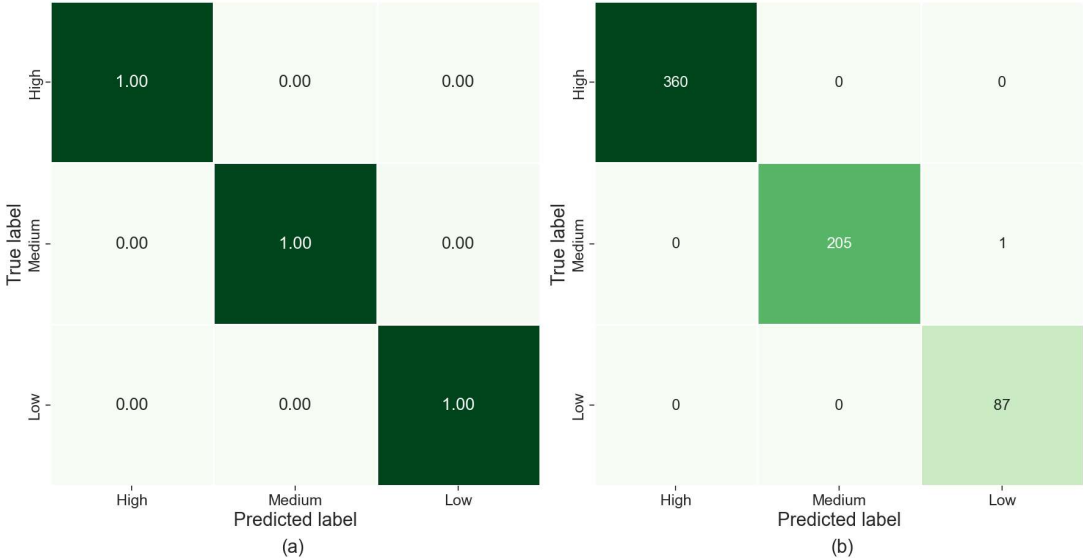

Figure 6. Confusion Matrix obtained by the validation dataset: (a) Normalised Confusion Matrix; (b) Confusion Matrix without Normalisation.

Figure 7 presents the ROC curve, illustrating the $TP$ and $FP$ rates as the decision threshold of the network varies. In this ROC curve, the diagonal line represents the performance of a random classifier, commonly referred to as the line of no discrimination. In Figure 7a, the red line indicates the performance of the "high" class, the orange line represents the "medium" class, and the green line corresponds to the "low" class. In Figure 7b the dashed line shows the overall performance of the model in terms of network efficiency. It should be noted that on the curve, all the classes evaluated individually coincide at the horizontal line because they each reach a $TP$ rate of 1.0. Additionally, the same Figure 7 displays the AUC values, which represent the area under the curve, with a value of 1 indicating perfect prediction. In this study, all three classes achieved an AUC of 1.0, obtained by rounding the actual network performance to four decimal places.

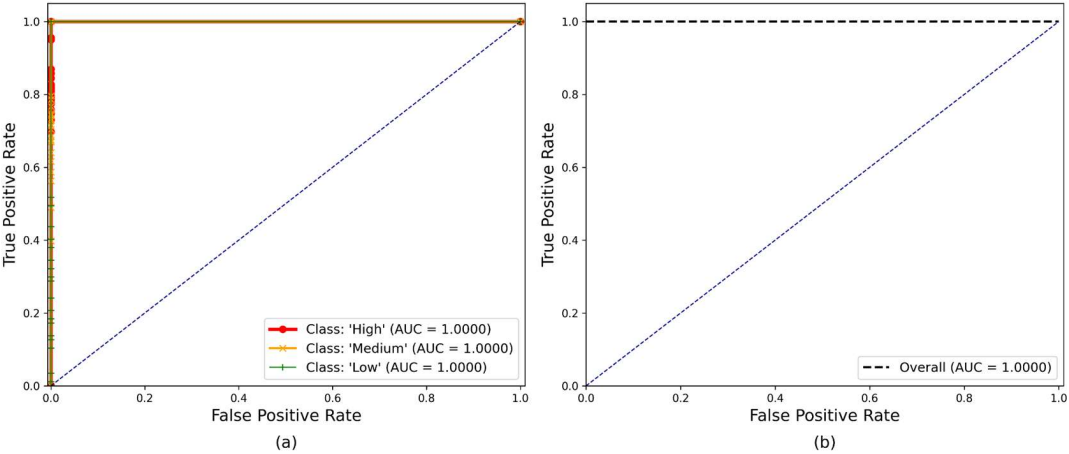

Figure 7. ROC curve obtained from the evaluation of CNN performance: (a) ROC curve and AUC value for each class; (b) ROC curve and overall AUC value.

Table 2 summarises the remaining evaluation metrics for the CNN performance on the training and validation datasets. This table includes the accuracy, precision, recall (or sensitivity), F-measure, and training time. The metrics for both training and validation data achieve a value of 0.9985. Notably, despite the relatively long training time, the CNN demonstrates excellent performance.

Table 2. Validation metrics obtained from the validation and training dataset.

| Metric | Validation values | Training values |
| --- | --- | --- |
| Accuracy | 0.9995 | 0.9997 |
| Precision | 0.9985 | 0.9997 |
| Recall | 0.9985 | 0.9997 |
| F-measure | 0.9986 | 0.9997 |
| Training time | - | 354 min 13.4 s |

Finally, we present a convolutional feature map of a random image from the validation dataset in Figure 8, which illustrates a region of interest detected by the CNN for classification purposes. Figure 8a shows the input image, which has been resized, normalised, and standardised for the CNN. Figure 8b represents a feature map obtained from the image after passing through convolutional layer 1, with dimensional reduction through max pooling. Figure 8c corresponds to the feature map extracted after convolutional layer 2, also with max pooling applied. Figures 8d, 8e, and 8f display the feature maps after passing through convolutional layers 3, 4, and 5, respectively. In this study, several feature maps from different randomly selected images were analysed to confirm that the CNN can detect changes in the image caused by suspended particles in liquid samples.

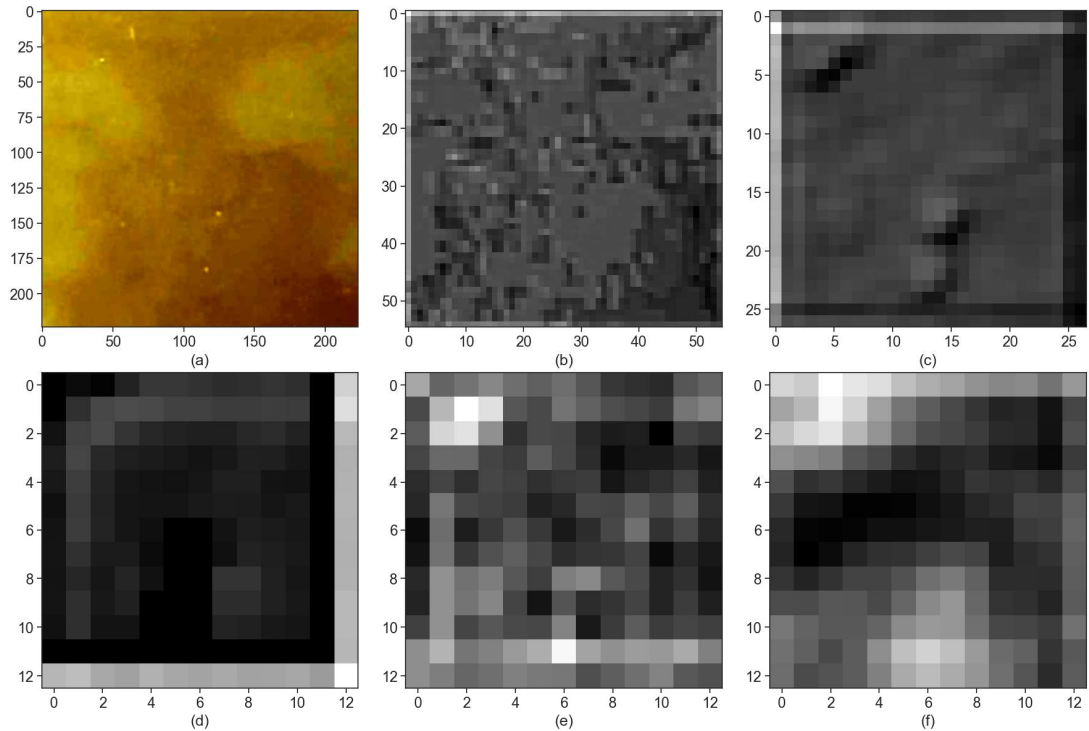

Figure 8. A random feature map from an image in the validation dataset through each convolutional layer. (a) Random input image; (b) Feature map after convolutional layer 1; (c) Feature map after convolutional layer 2; (d) Feature map after convolutional layer 3; (e) Feature map after convolutional layer 4; (f) Feature map after convolutional layer 5.

**DISCUSSION**

The issue of water pollution by TSS requires action from environmental policymakers concerning the connection of individuals to the public sewer system, as well as monitoring the health of water bodies. In this context, machine learning techniques offer valuable tools for decision-making based on physical data obtained from simple water imaging. The CNN developed in this study exemplifies a tool that delivers excellent performance with modest computational resources.

Figure 5, which depicts the learning curve of the proposed model, shows that the CNN developed for water quality classification based on TSS concentration demonstrates strong performance. This is evident from epoch 10 onwards, with an accuracy exceeding 0.99, along with high precision, sensitivity, and F1-measure. However, fluctuations occur in subsequent training epochs. These fluctuations are primarily due to images with low and medium solid concentrations, where the solid content is so minimal that it becomes challenging to identify within the sample, especially during constant agitation. With an agitation speed of 300 rpm and a selected image crop area of 450 x 450 pixels out of a total of 1920 x 1080 pixels, some images may not adequately capture the TSS concentration in the water sample. This crop area was chosen to avoid capturing water vortices and the magnetic stirrer used in the agitation process. The fluctuations in the learning process are likely due to difficulties in distinguishing between the low and medium classes. This is supported by the Confusion Matrix, which shows one misclassified image between the low class (label 3) and the medium class (label 2) in terms of true labels and predictions (Figure 6b). However, this misclassification occurs in only one of the 653 images used for validation, resulting in an error rate of 1/653, equivalent to an error of

0.0015. It is important to emphasise that overfitting can be ruled out based on the learning curve analysis. The learning curve shows that as the number of epochs increases, the model accuracy continues to improve. Although there are fluctuations in accuracy and loss, these variations are relatively small, within a range between 0.02 and 0.0004, especially when the accuracy reaches a high value of 0.99. This minor fluctuation indicates stability and high performance, suggesting that the model is learning effectively and is unlikely to be overfitting.

Regarding the "high" class in TSS concentration, it is evident that this class was the most accurately classified by the network. With the highest number of images used for this category, no misclassified labels were found, resulting in an accuracy of 1.0. This can be attributed to the higher TSS concentrations in these samples and the abundance of training data, enabling the proposed CNN model to correctly associate high concentrations with this label.

As shown by the previously presented metrics (Table 3), despite the presence of one misclassified image out of 653 used for validation, the ROC curve indicates that the classification probability is nearly 1.0 for all classes. Specifically, the "high" class achieves a perfect identification probability of 1.0. For the "medium" and "low" classes, the probability of correct identification is 0.99999, which is rounded to 1.0000 in ROC (Figure 7a) for clarity due to decimal precision. Additionally, the overall AUC for the model is 0.999996, also rounded to 1.0000 in ROC (Figure 7b), indicating a very high level of accuracy. These results suggest that the developed network has a high probability of accurately classifying TSS concentrations.

Regarding the feature maps presented in Figure 8, these provide a detailed view of how the network processes and classifies images, thereby validating its ability to detect relevant features. These feature maps reveal which features are activated at each layer of the network, illustrating how the network processes the image—from detecting simple edges and textures to identifying complex patterns. The review of these maps confirms that the CNN focuses on differences in suspended solids and effectively learns to identify relevant patterns, while ignoring factors such as optical distortions and unwanted radiation. Additionally, the dimensions of each feature map at the output of the layers demonstrate that the model is correctly structured according to the architecture used by AlexNet.

Although the developed CNN model demonstrates a very high classification probability, it is important to acknowledge its limitations. One such limitation pertains to the high class, where the samples used reached a concentration limit of 6000 mg/L, corresponding to sample 30. This value does not represent the maximum possible concentration of TSS, as concentrations exceeding 20,000 mg/L have been recorded, potentially leading to saturation of the water sample. In this study, only concentrations up to 6000 mg/L were characterised, thus limiting the CNN model to this concentration range. The larger number of samples in the "high" class was due to the potential for high TSS concentrations, given the greater variability in TSS within this water quality category.

Another important consideration is the impact of different lighting conditions. In this study, data capture was conducted in a controlled environment with constant lighting. However, varying lighting sources and capturing images from different angles could introduce classification errors in the CNN. Therefore, to maintain high predictive accuracy, it is recommended to continue using the experimental procedure employed for image collection.

Additionally, the proposed model is currently limited to estimating water quality classes in terms of TSS for samples without dissolved solids (DS). Dissolved solids can vary in nature—colloidal, organic, inorganic, or soluble materials—and may have different colorations, potentially causing classification errors. Therefore, this CNN is designed to assess water quality in contexts where high levels of dissolved solids are absent. Consequently, the model is suitable for evaluating water quality in drinking water distribution systems, treated wastewater, or rainwater, where high concentrations of dissolved solids are generally not present. Despite this limitation, the CNN model effectively distinguishes and classifies images of water with TSS concentrations, demonstrating its potential for TSS water quality classification, especially in waters with low levels of dissolved solids.

Another key feature of our CNN model is its reproducibility. To achieve this, we optimised the model design to require low computational costs. This is reflected in the training time of the network, which, despite lasting 354 minutes and 13.4 seconds, was executed on a computer with modest specifications, as we have already explained. The dataset used is relatively small compared to those employed in models handling more complex images. Additionally, the classifier has been specifically tuned to the characteristics of the image set and the available computing power, ensuring a balance between accuracy and efficiency. This strategy not only facilitates the replication of the model in other environments but also reduces the resources needed for its implementation.

Currently, determining TSS as a water pollution parameter is performed in water quality laboratories, which is both costly and time-consuming. Therefore, this study proposes a novel alternative for classifying solids. This approach could become a valuable tool, particularly in the environmental monitoring of water bodies, as it enables the evaluation of TSS in an easy, rapid, and efficient manner. By classifying TSS, it is possible to provide recommendations on the use, destination, and disposal of water. For instance, in Mexico, the "low" classification aligns with the TSS concentrations allowed for the discharge of treated wastewater into rivers, streams, canals, drains, reservoirs, lakes, and lagoons, and for irrigation of green areas. This class corresponds to water of acceptable quality in terms of TSS, while the "medium" class is associated with concentrations characteristic of contaminated water, and the "high" class with heavily contaminated water.

**CONCLUSIONS**

In this study, we developed a CNN framework utilising transfer learning to classify three levels of TSS in water. The CNN was trained using visual records of water samples with varying TSS concentrations. By employing TL, we were able to leverage pre-existing knowledge from large datasets, enabling our model to perform with near-perfect accuracy despite being trained on relatively small datasets. This approach not only demonstrates the feasibility of using TL in environmental monitoring but also highlights its effectiveness in achieving competitive classification performance comparable to models trained on much larger datasets. Our results suggest that TL can significantly reduce the need for extensive labelled data and computational resources, making it a promising tool for practical applications in water quality analysis and other environmental monitoring tasks.

The results archived from the validation of the proposed model demonstrate an adequate performance in predicting water quality classification based on the TSS parameter obtained from images. The main advantage of the model lies in the ease of acquiring these images, as they were captured with a smartphone and a Ring Light Edge-Lit LED lamp, a common lighting source for video recordings that is economically available. Nevertheless, despite the promising metrics achieved during the training and validation phases, further refinements are required to reach an accuracy of 1.0. With these enhancements and the appropriate model weights, the CNN has the potential to be effectively deployed for evaluating water quality in the context of detecting suspended solids. Achieving this level of performance will necessitate improvements to and expansion of the dataset, as well as enhancements in image collection, particularly for images with low concentrations of solids (low class). It can be theorised that reducing the rpm during sample agitation may yield a better dataset for training the CNN. Another potential improvement is the optimization of the network hyperparameters; however, this would require increased computational costs. So, in future work, we plan to develop a new model that classifies water into five quality levels: "excellent", "good", "acceptable", "contaminated", and "heavily contaminated", that is the classification used by the Ministry of Environment and Natural Resources Mexico to classify water quality by suspended solids. This will enhance the distribution of solids concentrations in the samples used in this study. Additionally, we aim to create a hybrid model that combines the proposed CNN with an ANN using Multiple Linear Regression (MLR) to estimate both contamination classification and TSS concentration in water samples. To achieve this, we will improve the current model by expanding the image dataset.

## ACKNOWLEDGEMENTS

We acknowledge Salomón Borjas García for valuable support.

## FUNDING

We acknowledge support from CIC-UMSNH under grants 18371 and 26140.

## AUTHOR CONTRIBUTIONS

All authors contributed equally to this research paper.

## DATA AVAILABILITY STATEMENT

GitHub will be available for dataset and code consultation.

## CONFLICT OF INTEREST

Authors declare no conflict of interest.

## ETHICS STATEMENT

**Ethics: Human participants**

No human participants were involved in this research.

**Ethics: Animal testing**

No animals were used in this research.